\documentclass[sigconf]{acmart}
\usepackage{algorithm}
\usepackage{algorithmic}
\usepackage{graphicx}
\usepackage{epstopdf}
\usepackage{subfigure}
\usepackage{soul}
\usepackage{multicol}
\usepackage{multirow}
\usepackage{float}
\usepackage{enumitem}
\usepackage{booktabs}
\usepackage[symbol]{footmisc}
\newcommand{\tabincell}[2]{\begin{tabular}{@{}#1@{}}#2\end{tabular}}

\AtBeginDocument{%
  \providecommand\BibTeX{{%
    \normalfont B\kern-0.5em{\scshape i\kern-0.25em b}\kern-0.8em\TeX}}}

\copyrightyear{2021}
\acmYear{2021}
\setcopyright{acmcopyright}\acmConference[WSDM '21]{Proceedings of the Fourteenth ACM International Conference on Web Search and Data Mining}{March 8--12, 2021}{Virtual Event, Israel}

\acmConference[WSDM '21] {Proceedings of the Fourteenth ACM International Conference on Web Search and Data Mining}{March 8--12, 2021}{Virtual Event, Israel}
\acmBooktitle{Proceedings of the Fourteenth ACM International Conference on Web Search and Data Mining (WSDM '21), March 8--12, 2021, Virtual Event, Israel}
\acmPrice{15.00}
\acmDOI{10.1145/3437963.3441782}
\acmISBN{978-1-4503-8297-7/21/03}

\begin{document}
\fancyhead{}
\title{Adversarial Immunization for Certifiable Robustness on Graphs}

\author{Shuchang Tao, Huawei Shen*, Qi Cao, Liang Hou, Xueqi Cheng}
\affiliation{
\institution{
CAS Key Lab of Network Data Science and Technology, 
Institute of Computing Technology, \\
Chinese Academy of Sciences
}
\country{
Beijing, China
}
}
\affiliation{
\institution{
University of Chinese Academy of Sciences
}
\country{
Beijing, China
}
}
\email{{taoshuchang18z,shenhuawei,caoqi,houliang17z,cxq}@ict.ac.cn}


\begin{abstract}
Despite achieving strong performance in semi-supervised node classification task, graph neural networks (GNNs) are vulnerable to adversarial attacks, similar to other deep learning models. Existing researches focus on developing either robust GNN models or attack detection methods against adversarial attacks on graphs. However, little research attention is paid to the potential and practice of immunization to adversarial attacks on graphs. In this paper, we propose and formulate the \emph{graph adversarial immunization} problem, i.e., vaccinating an affordable fraction of node pairs, connected or unconnected, to improve the certifiable robustness of graph against any admissible adversarial attack. We further propose an effective algorithm, called \emph{AdvImmune}, which optimizes with meta-gradient in a discrete way to circumvent the computationally expensive combinatorial optimization when solving the adversarial immunization problem. Experiments are conducted on two citation networks and one social network. Experimental results demonstrate that the proposed \emph{AdvImmune} method remarkably improves the ratio of robust nodes by 12$\%$, 42$\%$, 65$\%$, with an affordable immune budget of only 5$\%$ edges.
\let\thefootnote\relax\footnotetext{*Corresponding Author}
\end{abstract}



\maketitle

\section{Introduction}

Graph data are ubiquitous in the real world, characterizing complex relationships among objects or entities. 
Typical graph data include social networks, citation networks, biological networks, and traffic networks. 
In the last few years, graph neural networks (GNNs) emerge as a family of powerful tools to model graph data, achieving remarkable performance in many graph mining tasks such as node classification~\cite{kipf2017semi, Klicpera2018PredictTP, xu2018gwnn, xu2019graphheat}, cascade prediction~\cite{Cao2020PopularityPO,Qiu2018DeepInfSI}, and recommendation systems~\cite{fan2019graph}.
Despite their success, similar to deep learning models in other fields~\cite{Carlini2016TowardsET,Jin2019IsBR}, GNNs are proved to be vulnerable to adversarial attacks~\cite{Dai2018AdversarialAO,zugner2018adversarial,Bojchevski2018AdversarialAO}, i.e., imperceptible perturbations on graph structure or node features can easily fool GNNs. 

\begin{figure}
\centering
\includegraphics[width=7cm]{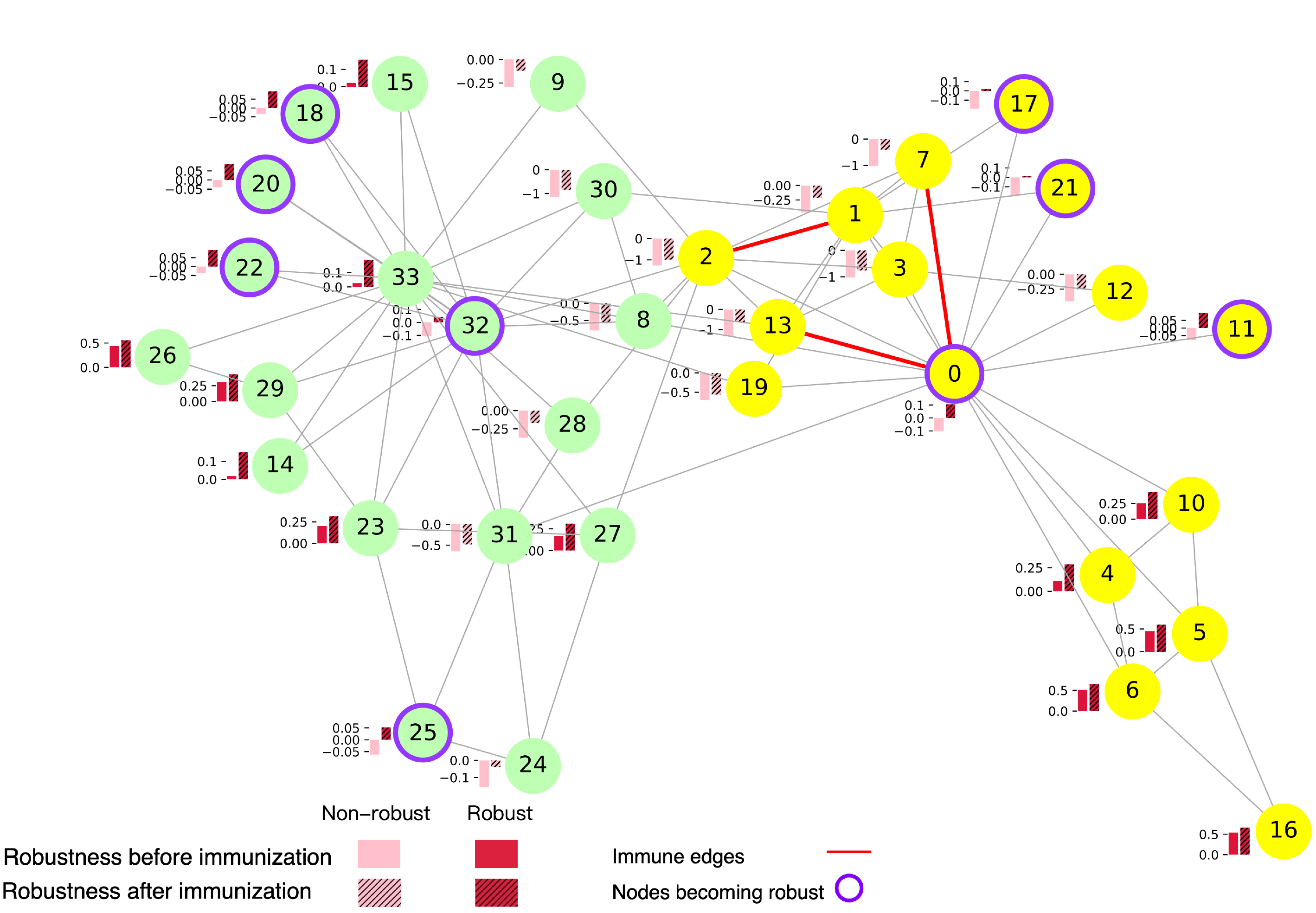}
\caption{Effect of adversarial immunization on Karate club network. 
Colors differentiate nodes in two classes.  
We use two bars to represent node's robustness before and after immunization. 
The node is certified as robust (red), when its robustness $>$ 0, otherwise as non-robust (pink).
	Purple circle indicates the node that becomes robust through immunization.
	The red edges are immune edges.
}
\label{fig:karate}
\end{figure}

Many researchers devote to designing defense methods against adversarial attacks to GNN models, such as adversarial training~\cite{feng2019graph,Dai2019AdversarialTM} and attack detection~\cite{zhang2019comparing}. 
Defense methods spring up rapidly and gain success at improving the performance of GNN models
~\cite{Wu2019AdversarialEF,Xu2019TopologyAA,Zhu2019RobustGC}. However, these methods are usually heuristic~\cite{Bojchevski2019CertifiableRT} and only effective for certain attacks rather than all attacks. 
Consequently, an endless cat-and-mouse game or competition emerges between adversarial attacks and defense methods~\cite{Bojchevski2019CertifiableRT}. 
Recently, to solve the attack-defense dilemma, researchers resort to robustness certification and robust training~\cite{zugner2020CertiRob,Liu2020CertifiableRT} on graphs against any admissible adversarial attack~\cite{Bojchevski2019CertifiableRT,zugner_adversarial_2019}. However, such robust training may damage the performance of GNN models on clean graph, which is undesirable before adversarial attack actually happens. 

Up to now, little research attention is paid to the potential and practice of immunization to adversarial attacks on graphs. 
In this paper, we firstly propose and formulate  \emph{graph adversarial immunization}, which is the first action guideline to improve the certifiable robustness against any admissible adversarial attack from the perspective of graph structure instead of GNN models or training methods. Specifically, \emph{adversarial immunization vaccinates a fraction of node pairs in advance, connected or unconnected, to protect them from being modified by attacks,} making the whole graph more robust to adversarial attacks.  
Note that adversarial immunization is general and flexible, not only improving certifiable robustness against any admissible attack, but also avoiding the cons of performance drop on unattacked clean graph suffered by robust training~\cite{Bojchevski2019CertifiableRT}.
We further propose an effective algorithm called \textit{AdvImmune} to obtain the target immune node pairs with meta-gradient in a discrete way, circumventing the computationally expensive combinatorial optimization when solving 
 adversarial immunization. 

To offer an intuitive understanding about adversarial immunization, we illustrate its effect on an example network, i.e., the Karate club network. 
As shown in  Figure~\ref{fig:karate}, only immunizing 3 edges in advance, the number of nodes certified as robust against any admissible attack increases by 9. 
Indeed, the remarkable improvement of certifiable robustness is also observed when applying our proposed \textit{AdvImmune} method on real large-scale networks. 
Experimental results on Reddit show that with an affordable immune budget of only 5\% edges, the ratio of robust nodes improves by 65\%. 
Such results indicate the effectiveness of our proposed \textit{AdvImmune} method, i.e., immunizing certain affordable critical node pairs in advance can significantly improve the certifiable robustness of graph.

Adversarial immunization has much potential in real world application. For example, in recommendation systems, immunization protects users from junk or low-quality products by maintaining the relationship between real users and products; in credit scoring systems~\cite{Jin2020AdversarialAA}, immunization can maintain certain critical relation of user pairs to prevent fraudsters from pretending to be normal customers, avoiding serious financial losses.

In summary, our contributions are as follows:
 \begin{enumerate}[topsep = 0.5 em]
\item To the best of our knowledge, this is the first work that proposes and formalizes \emph{adversarial immunization on graph} for certifiable robustness against any admissible attack, providing a brand new insight into graph adversarial learning.
\item We propose \textit{AdvImmune} algorithm to innovatively tackle adversarial immunization with meta-gradient to obtain the optimal immune node pairs, circumventing the computationally expensive combinatorial optimization.
\item We demonstrate the effectiveness of our \textit{AdvImmune} method on two citation networks and one social network. Experimental results show that our proposed method significantly improves the ratio of robust nodes by 12\%, 42\% and 65\%, with an affordable immune budget of only 5\% edges.
\end{enumerate}

\section{Preliminaries}
\label{sec:Pre}
Since adversarial learning on graphs mainly take semi-supervised node classification as the target task, this section first introduces the task of semi-supervised node classification, together with a widely used GNN model to tackle this task.
Besides, we also introduce the robustness certification against any admissible attack. 

\textbf{Semi-supervised node classification.}
Given an attributed gra-ph $G=(\boldsymbol{A}, \boldsymbol{X})$, $ \boldsymbol{A} \in\{0,1\}^{N \times N} $ is the adjacency matrix and $\boldsymbol{X} \in \mathbb{R}^{N \times d}$ is the attribute matrix consisting of node attributes, $N$ is the number of nodes and $d$ is the dimension of node attributes. 
We denote the node set as $\mathcal{V} =\{1, 2, ..., N\}$ and the edge set as $\mathcal{E} \subseteq \mathcal{V} \times \mathcal{V}$.
In semi-supervised node classification,  a subset of nodes $\mathcal{V}_l \in \mathcal{V}$ are labelled from class sets $\mathcal{K}$.
The goal is to assign a label for each unlabelled node by the learned classifier $\boldsymbol{Y}=f(\boldsymbol{A}, \boldsymbol{X}) \in \mathbb{R}^{N \times K}$, where $K=|\mathcal{K}|$ is the number of classes.

\textbf{Graph neural networks.}
GNNs have achieved a remarkable success on semi-supervised node classification task~\cite{kipf2017semi, velickovic2018graph, huang2018adapt, Klicpera2018PredictTP}. 
Among existing GNN models, $\pi$-PPNP~\cite{Klicpera2018PredictTP,Bojchevski2019CertifiableRT} shows an outstanding performance.
It connects graph convolutional networks (GCN) to PageRank to effectively capture the impact of infinitely neighborhood aggregation, and decouples the feature transformation from propagation to simplify model structure.
In this paper, we consider $\pi$-PPNP as the representative of GNN models to tackle node classification problem.
The formulation is as follows:
\begin{equation}
	\begin{aligned}
	\boldsymbol{H} &=f_{\theta}\left(\boldsymbol{X}\right) \\
    \boldsymbol{Y} &=\operatorname{softmax}\left(\boldsymbol{\Pi}\boldsymbol{H}\right),
    \label{eq:PPNP}
\end{aligned}
\end{equation}
where $\boldsymbol{H} \in \mathbb{R}^{N \times K}$ is the transformed features computed by a neural network $f_{\theta}$, $\boldsymbol{H}^{\text {diff }}:=\boldsymbol{\Pi} \boldsymbol{H}$ is defined as the \textit{diffused logits}. Note that the diffused logits are the unnormalized outputs with real numbers ranging from $(-\infty,+\infty)$, which are also referred to raw predictions. $\mathbf{\Pi} =(1-\alpha)\left(\boldsymbol{I}_{N}-\alpha \boldsymbol{D}^{-1} \boldsymbol{A} \right)^{-1}$ is personalized PageRank~\cite{Page1999ThePC} that measures distance between nodes, and $\boldsymbol{D}$ is a diagonal degree matrix with $\boldsymbol{D}_{i i}=\sum_{j} \boldsymbol{A}_{i j}$. 
The personalized PageRank  with source node $t$ on graph $G$ is written as: 
\begin{equation}
	\boldsymbol{\pi}_G\left(\boldsymbol{e}_{t}\right)= (1-\alpha) \boldsymbol{e}_{t}\left(\boldsymbol{I}_{N}-\alpha \boldsymbol{D}^{-1} \boldsymbol{A}\right)^{-1}, 
	\label{eq:ppr}
\end{equation}
where $\boldsymbol{e}_t$ is the $t$-th canonical basis vector (row vector).  
$\boldsymbol{\pi}_G\left(\boldsymbol{e}_{t}\right)=\mathbf{\Pi}_{t,:}$ is the $t$-th row of personalized PageRank matrix $\mathbf{\Pi}$.

\textbf{Robustness certification.} 
\label{sec:graph_cert}
Since GNN models are vulnerable to adversarial attacks, 
Bojchevski \textit{et al.} ~\cite{Bojchevski2019CertifiableRT} certify the robustness of each node against any admissible attack on graph by \textit{worst-case margin}.
Specifically, the difference between the raw predictions of node $t$ on label class $y_t$ and other class $k$ defines the \textit{margin} on the perturbed graph $\tilde{G}$. Taking $\pi$-PPNP as a typical GNN model, the formula of the defined margin is as follows:
 \begin{equation}
 	 \mathbf{m}_{y_t, k}(t,\tilde{G})
 	 =\boldsymbol{H}_{t, y_t}^{\text {diff}}-\boldsymbol{H}_{t, k}^{\mathrm{diff}}
 	 =\boldsymbol{\pi}_{\tilde{G}}\left(\boldsymbol{e}_{t}\right)\left(\boldsymbol{H}_{:, y_t}-\boldsymbol{H}_{:, k}\right),
 	  	\label{eq:w-margin cert}
 \end{equation}
where $y_t$ is the label class of node $t$.

For a target node $t$, the \textit{worst-case margin} between class $y_t$ and class $k$ under any admissible perturbation $\tilde{G} \in \mathcal{Q}_{\mathcal{F}}$ is:
\begin{equation}
\mathbf{m}_{y_{t}, k}(t,\tilde{G}^{*})=\min _{\tilde{G} \in \mathcal{Q}_{\mathcal{F}}} \mathbf{m}_{y_{t}, k}(t,\tilde{G})
=\min _{\tilde{G} \in \mathcal{Q}_{\mathcal{F}}} \boldsymbol{\pi}_{\tilde{G}}\left(\boldsymbol{e}_{t}\right)\left(\boldsymbol{H}_{:, y_{t}}-\boldsymbol{H}_{:, k}\right),
\label{eq:PPNP_cert}
\end{equation}
where $\tilde{G}=(\mathcal{V}, \tilde{\mathcal{E}}:=\mathcal{E}_{f} \cup \mathcal{F}_{+})$ is the perturbed graph consisting of fixed edges $\mathcal{E}_{f}$ and the perturbed edges $\mathcal{F}_{+}$, $\tilde{G}^{*}$ is the worst-case perturbed graph.
The set of admissible perturbed graphs
$
\mathcal{Q}_{\mathcal{F}}=\left\{\left(\mathcal{V}, \tilde{\mathcal{E}}:=\mathcal{E}_{f} \cup \mathcal{F}_{+}\right) | 
\mathcal{F}_{+} \subseteq \mathcal{F}, 
\left|\tilde{\mathcal{E}} \backslash \mathcal{E}\right|+\left|\mathcal{E} \backslash \tilde{\mathcal{E}}\right| \leq B,
\left|\tilde{\mathcal{E}}^{t} \backslash \mathcal{E}^{t}\right|+\right.\\
\left.\left|\mathcal{E}^{t} \backslash \tilde{\mathcal{E}}^{t}\right|
 \leq b_{t}, \forall t\right\}
$
, where $\mathcal{F}\subseteq(\mathcal{V} \times \mathcal{V}) \backslash \mathcal{E}_{f}$ is fragile edge set.
Each fragile edge $(i, j) \in \mathcal{F}$ can be included in the graph or excluded from the graph by attacker, and the selected attack edge $\mathcal{F}_{+}$ is a subset of $\mathcal{F}$. 
The perturbations satisfy both global budget and local budget, where global budget $|\tilde{\mathcal{E}} \backslash \mathcal{E}|+|\mathcal{E} \backslash \tilde{\mathcal{E}}| \leq B$ requires that there are at most $B$ perturbing edges, and local budget $|\tilde{\mathcal{E}}^{t} \backslash \mathcal{E}^{t}|+| \mathcal{E}^{t} \backslash \tilde{\mathcal{E}}^{t} | \leq b_{t}$ limits node $t$ to have no more than $b_t$ perturbing edges.

Node $t$ is certifiably robust when: 
\begin{equation}
	\mathbf{m}_{y_{t}, k_t}(t,\tilde{G}^{*})=\min _{k \neq y_{t}} \mathbf{m}_{y_{t}, k}(t,\tilde{G}^{*})>0,
	\label{eq:min w-margin}
\end{equation}
where $k_t$ is the most likely class among other classes.
In other words, whether a node is robust is determined by the worst-case margin against any admissible perturbed graph. 
This certification method directly measures the robustness of node/graph under GNNs. 

Bojchevski \textit{et al.}~\cite{Bojchevski2019CertifiableRT} use policy iteration with reward $\boldsymbol{r}=-\left(\boldsymbol{H}_{:, y_t}\right.\\ \left.-\boldsymbol{H}_{:, k}\right)$ to find the worst-case perturbed graph. 
Under certain set of admissible perturbed graph $\mathcal{Q}_\mathcal{F}$, running policy iteration $K \times (K-1)$ times can obtain the certificates for all $N$ nodes.

\section{Adversarial Immunization}
In this section, we first formalize the problem of graph adversarial immunization and elaborate it with a widely used GNN model. 
Then we propose an effective algorithm \textit{AdvImmune}, using meta-gradient for selecting and immunizing appropriate node pairs in advance, to improve the certifiable robustness of graph.

\subsection{Problem Formulation}
Adversarial immunization aims to improve the certifiable robustness of nodes against any admissible attack, i.e., the minimal \textit{worst-case} margin of nodes under node classification task. Specifically, by immunizing appropriate node pairs in advance, GNN model can correctly classify nodes even under the worst case. In this paper, we use $\pi$-PPNP as the typical GNN model, since it shows an outstanding performance on node classification task. The general goal of adversarial immunization is formalized as:
\begin{equation}
\max _{\mathcal{E}_{\boldsymbol{c}}\in \mathcal{S}_{\boldsymbol{c}}}  \min _{k \neq y_{t}}  \mathbf{m}_{ y_{t}, k}(t, \hat{G})=
\max _{\mathcal{E}_{\boldsymbol{c}} \in \mathcal{S}_{\boldsymbol{c}}} \min _{k \neq y_{t}} \boldsymbol{\pi}_{\hat{G}}\left(\boldsymbol{e}_{t}\right)\left(\boldsymbol{H}_{:, y_{t}}-\boldsymbol{H}_{:, k}\right),
\label{eq:node_goal}
\end{equation}
where $\boldsymbol{\pi}_{\hat{G}}$ is the personalized PageRank of the modified graph, $\hat{G} = \left(\mathcal{V}, \hat{\mathcal{E}} :=\left(\mathcal{E}_{\tilde{G}^{*}} \cup \mathcal{E}_{\boldsymbol{c}}^{connect}\right) \backslash \mathcal{E}_{\boldsymbol{c}}^{unconnect}\right)$ is the modified graph with contribution of both perturbed graph $\mathcal{E}_{\tilde{G}^{*}}$ and immune graph $\mathcal{E}_{\boldsymbol{c}}=\left(\mathcal{E}_{\boldsymbol{c}}^{connect} \cup \mathcal{E}_{\boldsymbol{c}}^{unconnect}\right) \subseteq (\mathcal{V} \times \mathcal{V})$,  
$\mathcal{E}_{\tilde{G}^{*}}$ is the edge set of the worst-case perturbed graph $\tilde{G}^{*}$: 
\begin{equation}
\tilde{G}^{*} = \underset{\tilde{G}\in \mathcal{Q}_{\mathcal{F}}}{\arg \min } \ \mathbf{m}_{ y_{t}, k}(t, \tilde{G}),
\end{equation}
and the immune graph  $\mathcal{E}_{\boldsymbol{c}}$ contains both connected edges $\mathcal{E}_{\boldsymbol{c}}^{connect}$ to keep them in the graph and unconnected node pairs $\mathcal{E}_{\boldsymbol{c}}^{unconnect}$ to keep them not in the graph. 
Due to the limited immune budget in reality, we cannot immunize all node pairs. 
Here, we consider both local budget and global budget to constrain the choice of immune node pairs $\mathcal{E}_{\boldsymbol{c}}$. 
Globally, the number of immune node pairs should be no more than \textit{global budget} $C$, i.e., $\left|\mathcal{E}_{\boldsymbol{c}}\right| \leq C$.
For each node $t$, the number of immune node pairs cannot exceed the \textit{local budget} $c_t$, i.e., $\left|\mathcal{E}_{\boldsymbol{c}}^{t}\right| \leq c_t$, where $\mathcal{E}_{\boldsymbol{c}}^{t}$ represents the node pairs associated with node $t$. 
The set of admissible immune node pairs $\mathcal{E}_{\boldsymbol{c}}$ is defined as:
\begin{equation}
\mathcal{S}_{\boldsymbol{c}} = \left\{ \mathcal{E}_{\boldsymbol{c}}| \mathcal{E}_{\boldsymbol{c}} \subseteq (\mathcal{V} \times \mathcal{V}),
\left|\mathcal{E}_{\boldsymbol{c}}\right| \leq C, 
\left|\mathcal{E}_{\boldsymbol{c}}^{t}\right|\leq c_t,\forall t \in \mathcal{V} \right\}.
\end{equation}

Note that the above immunization objective function, i.e., Ep.~\ref{eq:node_goal} is for a single node $t$. 
In order to improve certifiable robustness of the entire graph, we take sum of the worst-case margins over all nodes as the overall goal of adversarial immunization:
\begin{equation}
	\max _{\mathcal{E}_{\boldsymbol{c}}\in \mathcal{S}_{\boldsymbol{c}}} \sum _{t \in \mathcal{V} }  \min _{k \neq y_{t}}  \boldsymbol{\pi}_{\hat{G}}\left(\boldsymbol{e}_{t}\right)\left(\boldsymbol{H}_{:, y_{t}}-\boldsymbol{H}_{:, k}\right).
	\label{eq:goal}
\end{equation}

\textbf{Challenges:} It's not easy to obtain the optimal immune node pairs due to two issues.
First, the computational cost on selecting the set of certain node pairs from the total node pairs is expensive. Given a global immune budget of $C$, possible immunization plans are $\left(\begin{array}{c}N^{2} \\ C\end{array}\right)$. This leads to an unbearable search cost $\mathcal{O}\left(N^{2 C}\right)$, making it difficult to find the optimal immune node pairs efficiently. 
The second challenge comes from the discrete nature of graph data. The resulting non-differentiability hinders back-propagating gradients to guide the optimization of immune node pairs.

\subsection{AdvImmune, an Effective Algorithm for Adversarial Immunization}
We propose \textit{AdvImmune} to greedily obtain the solution via meta-gradient, addressing the graph adversarial immunization effectively.
To facilitate the solution of meta-gradient, we regard immune node pairs as hyperparameters in matrix form.

\textbf{Matrix form of the problem.}
We use adjacency matrix to represent the set of immune node pairs in Eq.~\ref{eq:goal}
and formalize the optimization goal with matrix form:
\begin{equation}
	 \max _{\mathcal{A}_{\boldsymbol{c}}\in \mathcal{A}_{\mathcal{S}_{\boldsymbol{c}}}} \sum _{t \in \mathcal{V} } \min _{k \neq y_{t}} \boldsymbol{\pi}_{\hat{G}}\left(\boldsymbol{e}_{t}\right)\left(\boldsymbol{H}_{:, y_{t}}-\boldsymbol{H}_{:, k}\right),
	 \label{eq:goal_mat}
\end{equation}	 
\begin{equation*}
	\boldsymbol{\pi}_{\hat{G}}\left(\boldsymbol{e}_{t}\right)=(1-\alpha)\boldsymbol{e}_{t}\left(\boldsymbol{I}_{N}-\alpha \boldsymbol{D}^{-1}_{\hat{G}} \boldsymbol{A}_{\hat{G}}\right)^{-1}, \quad \boldsymbol{A}_{\hat{G}} = \boldsymbol{A}+\boldsymbol{A}^{\prime}_{\tilde{G}^{*}} * \mathcal{A}_{\boldsymbol{c}},
\end{equation*}
where $\mathcal{A}_{\boldsymbol{c}}$ is the matrix form of immune graph $\mathcal{E}_{\boldsymbol{c}}$, $\boldsymbol{A}^{\prime}_{\tilde{G}^{*}}$ is the matrix form of perturbing edge set corresponding to the worst-case perturbed graph $\tilde{G}^{*}$, $\boldsymbol{A}_{\hat{G}}$ is the adjacency matrix of modified graph $\hat{G}$ with the contribution of both worst-case perturbing graph $\boldsymbol{A}^{\prime}_{\tilde{G}^{*}}$ 
and immune graph $\mathcal{A}_{\boldsymbol{c}}$, and $*$ is element-wise multiplication. 
In immune graph $\mathcal{A}_{\boldsymbol{c}}$, $0$ indicates that the corresponding node pair will be immunized, which will filter the influence of perturbations, while $1$ implies that the corresponding node pair is not immunized and may be attacked.  
In other words,  $\mathcal{A}_{\boldsymbol{c}}$ can be regarded as a mask, protecting these immune node pairs from being modified or attacked. Such a matrix form transforms the original computationally expensive combinatorial optimization of node pair set into a discrete matrix optimization problem, reducing difficulty of obtaining the optimal solution. 

The next key problem is how to solve the optimization problem with matrix form.
We innovatively tackle it by  greedy algorithm via meta-gradient to obtain the optimal immune graph matrix.

\textbf{Meta-gradient of immune graph matrix.}
Meta-gradient is referred to as the gradient of hyperparameters~\cite{Finn2017ModelAgnosticMF, zugner_adversarial_2019}. 
Regarding immune graph matrix $\mathcal{A}_{\boldsymbol{c}}$ as a hyperparameter, we can calculate the meta-gradient of $\mathcal{A}_{\boldsymbol{c}}$ with the objective function:
\begin{equation}
\nabla_{\mathcal{A}_{\boldsymbol{c}}}^{\mathrm{meta}}
=\nabla_{\mathcal{A}_{\boldsymbol{c}}}
\left[\sum _{t \in \mathcal{V}} \min_{k \neq y_{t}} \boldsymbol{\pi}_{\hat{G}}\left(\boldsymbol{e}_{t}\right)\left(\boldsymbol{H}_{:, y_{t}}-\boldsymbol{H}_{:, k}\right)\right],
\label{eq:meta}
\end{equation}
where $\nabla_{\mathcal{A}_{\boldsymbol{c}}}^{\mathrm{meta}}$ is the meta-gradient of immune graph matrix $\mathcal{A}_{\boldsymbol{c}}$.
Each entry in $\nabla_{\mathcal{A}_{\boldsymbol{c}}}^{\mathrm{meta}}$ represents the impact of the corresponding node pair on the objective function, i.e., worst-case margin. 
Note that before computing meta-gradient, we should first calculate the most likely class $k_t$ which minimizes the \textit{worst-case} margin of each node $t$: 
\begin{equation}
	k_t = \underset{k\neq y_t }{\arg \min } \ \boldsymbol{\pi}_{\hat{G}}\left(\boldsymbol{e}_{t}\right)\left(\boldsymbol{H}_{:, y_{t}}-\boldsymbol{H}_{:, k}\right).
	\label{eq:k}
\end{equation}

Directly performing gradient optimization may result in decimals and elements greater than 1 or smaller than 0 in immune graph  matrix, making $\mathcal{A}_{\boldsymbol{c}}$ no longer a matrix indicating the immune choice of node pairs in graph. To solve this challenge, we optimize the immune graph matrix in a discrete way by greedy algorithm. 

\textbf{Greedy immunization via meta-gradient.}
Since the objective in immunization is a maximization problem, we apply discrete gradient ascent to solve it. In other words, we calculate the meta-gradient for each entry of the matrix $\mathcal{A}_{\boldsymbol{c}}$, and choose node pairs with the greatest impact greedily. 
Since $\mathcal{A}_{\boldsymbol{c}}$ is initialized as a matrix with all elements of 1, it can only be changed from 1 to 0. 
Hence, only the negative gradient is helpful. 
We define the inverse of meta-gradient as the value of the corresponding node pair:
$
	V_{(i,j)} = -\nabla_{\mathcal{A}_{c_{(i,j)}}}^{\mathrm{meta}}
$, which represents the impact of the corresponding node pair on the goal of adversarial immunization.  
 
 In order to preserve node pairs with the greatest impact, we greedily select entries with the maximum value in $V$:
 \begin{equation}
 	 (i^*,j^*) =\underset{(i,j)}{\arg \max } \  V_{(i,j)},
 \end{equation}
 and set the corresponding entries in $\mathcal{A}_{\boldsymbol{c}}$ to zero as immune node pairs, protecting them from being attacked and modified.

\begin{algorithm}[tp]
\caption{\textit{AdvImmune} immunization on graphs}
\label{alg:A}
\begin{algorithmic}[1]
	\STATE \textbf{Input:} Graph $G=(\boldsymbol{A}, \boldsymbol{X})$, \textit{immune budget}  $(C,c)$, 
	\STATE \textbf{Output:} Immune graph matrix $\mathcal{A}_{\boldsymbol{c}}$
	\STATE  $\tilde{G}^{*} \leftarrow$ train surrogate model to get worst-case graph
	\STATE Initialize the immune graph matrix as $\mathcal{A}_{\boldsymbol{c}} = ones(N,N)$
	\STATE \textbf{while} number of immune node pairs in $\mathcal{A}_{\boldsymbol{c}}$  $< C$  \textbf{do}
	\STATE \quad $k_t \leftarrow $ the class which minimizes the worst-case margin of each node $t$ as Eq.~\ref{eq:k}
	\STATE \quad $\hat{G} \leftarrow$ after perturbing with $\tilde{G}^*$ of the worst-case class $k_t$ and immunizing with $\mathcal{A}_c$.
	\STATE \quad $\nabla_{\mathcal{A}_{\boldsymbol{c}}}^{\mathrm{meta}}
\leftarrow $ the meta-gradient computing in Eq.~\ref{eq:meta}
	\STATE \quad Value $ V_{(i,j)} \leftarrow -\nabla_{a_{(i,j)}}^{\mathrm{meta}}$ 
	\STATE \quad Select the node pairs that have already been immunized and set the corresponding entries to 0 in $V$. 
	\STATE \quad $(i^*,j^*)\leftarrow $ maximal entry in $V$ which satisfies $c$.
	\STATE \quad $\mathcal{A}_{\boldsymbol{c}}[i^*,j^*] = 0 \leftarrow $ immunize one more node pair $(i^*,j^*)$.
	\STATE \textbf{end}
	\STATE \textbf{Return: } $\mathcal{A}_{\boldsymbol{c}}$
\end{algorithmic}
\end{algorithm}

\subsection{Algorithm}

\begin{figure}[t]
	\includegraphics[width=6.7cm]{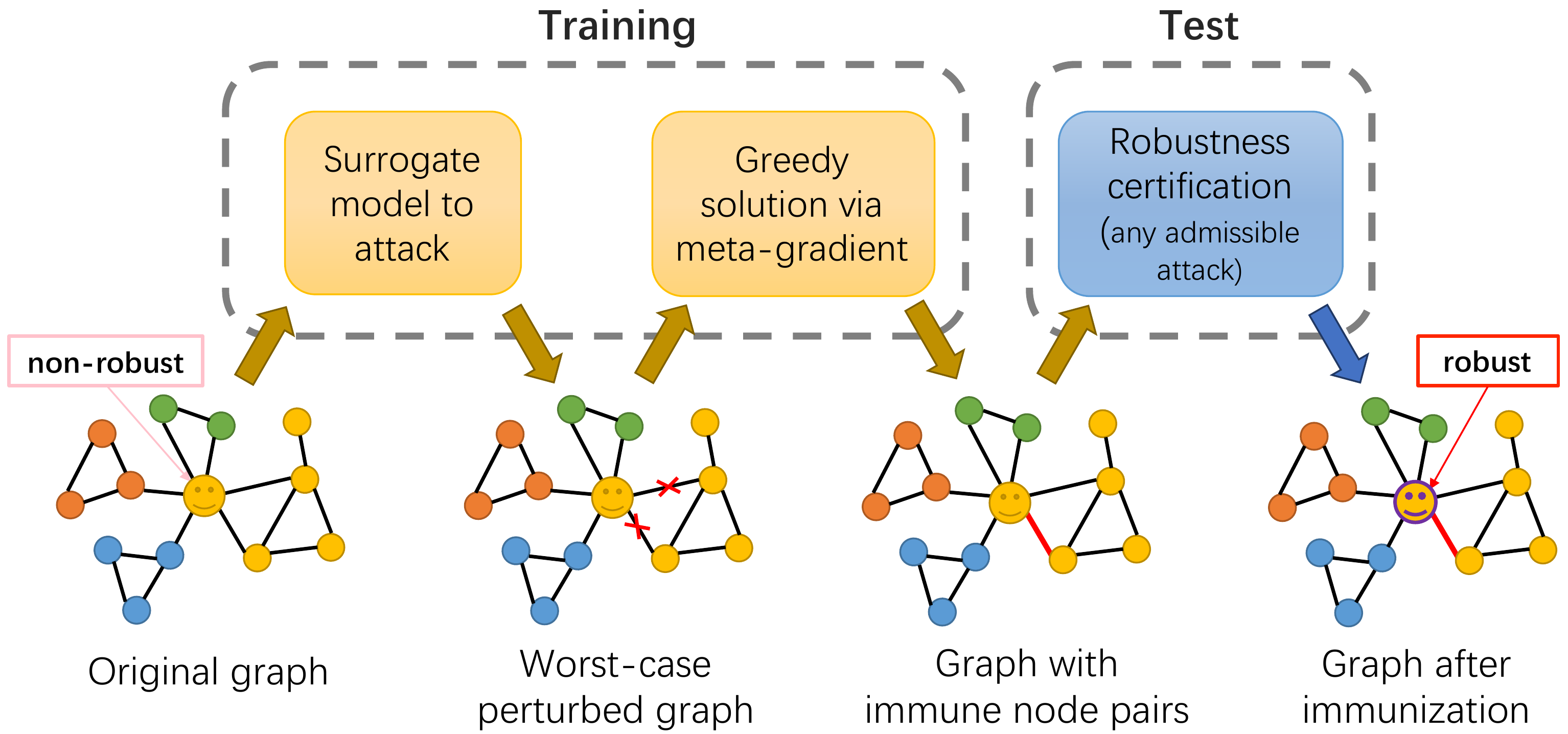}
	\caption{The training and test process of \textit{AdvImmune}. }
	\label{fig:process}
\end{figure}

In this section, we describe the \textit{AdvImmune} algorithm to obtain immune node pairs to improve the certifiable robustness of graph (Algorithm \ref{alg:A}).
We first initialize $\mathcal{A}_{\boldsymbol{c}}$ as a matrix with all elements of 1, 
indicating that no node pair is immunized.
Then we iteratively select node pairs with the greatest impact. 
In each iteration, we choose the most likely class $k_t$ which minimizes the worst-case margin of node $t$. 
Then the  core step is to calculate the meta-gradient of the objective function for  $\mathcal{A}_{\boldsymbol{c}}$ and the corresponding value $V_{(i,j)}$.
We select the maximum entry in $V_{(i,j)}$ which satisfies \textit{local budget} $c_t$, and set the corresponding entry in $\mathcal{A}_{\boldsymbol{c}}$ to zero. 
This process is repeated until we obtain enough immune node pairs.

We illustrate both training and test process of \textit{AdvImmune} in Figure~\ref{fig:process}.  
During the training process, we first use surrogate model to obtain the worst-case perturbed graph. 
Then, we select and immunize appropriate node pairs through Algorithm~\ref{alg:A} via meta-gradient.
As for test, we use the immune graph as a mask to protect certain node pairs from being modified, and improve the certifiable robustness of nodes against any admissible adversarial attack.

\subsection{Complexity Analysis}
The computational complexity of \textit{AdvImmune} algorithm depends on GNN model. 
Suppose the computational complexity of GNN model, i.e., the complexity of the surrogate attack model, is $\mathcal{O}(T)$. Specifically, under $\pi$-PPNP, the personalized PageRank matrix $\boldsymbol{\pi}$ is dense, leading to a computational complexity and memory requirement of  $\mathcal{O}(T) = \mathcal{O}(N^{2})$. 
Since we have to calculate the minimal worst-case margins for each pair of classes $(k_1, k_2)$, the computational complexity in each iteration is $\mathcal{O}\left( K^{2}\cdot T \right) $.
As for the optimization process of immunization, operations of both element-wise multiplication and meta-gradient require a computational complexity and memory of $\mathcal{O}\left(E\right)$.
Considering the total of $C$ iterations to find the optimal $C$ node pairs to be immunized, \textit{AdvImmune} method has a computational complexity of $\mathcal{O}\left(C \cdot K^{2}\cdot (T+E)\right)$.

\section{Experiments}
\subsection{Datasets}
We evaluate our proposed method on two commonly used citation networks: Citeseer and  Cora-ML~\cite{Bojchevski2019CertifiableRT}, and a social network Reddit~\cite{Zeng2019GraphSAINTGS}. In citation networks, a node represents a paper with  key words as attributes and paper class as label, and the edge represents the citation relationship.  
In Reddit, each node represents a post, with word vectors as attributes and community as label, while each edge represents the post-to-post relationship.

Due to the high complexity of $\pi$-PPNP, it is difficult to apply it to large graphs, leading to the limitation of our method.
Therefore, we only keep a subgraph of Reddit to conduct experiments. 
Specifically, we first randomly select 10,000 nodes, and four classes with the most nodes are selected as our target classes.
All nodes in these target classes are kept.
Then, in order to retain the network structure as much as possible, we further include the first-order neighbors of the kept nodes.
Nodes that don't belong to the target four classes is marked as the fifth class, i.e., other-class.

Experiments are conducted on the largest connected component (LCC) for both citation networks and social network. The statistics of each dataset are summarized in Table \ref{tab:dataset}. 

 \begin{table}[t]
  \caption{Statistics of the evaluation datasets}
  \label{tab:dataset}
  \begin{tabular}{ccccccc}
    \toprule
    Dataset & Type & $N_{LCC}$ & $\left|\mathcal{E}_{LCC}\right|$  & $d$ & $K$ \\
    \midrule
    Citeseer & Citation network & 2110 & 3668 & 3703 & 6\\
    Cora-ML & Citation network & 2810 & 7981  & 2879 & 7\\
    Reddit & Social network & 3069 & 7009 & 602 & 5 \\
  \bottomrule
\end{tabular}
\end{table}

\subsection{Baselines}
\label{sec:baseline}
To the best of our knowledge, we are the first to propose adversarial immunization on graphs. 
To demonstrate the effectiveness of our \textit{AdvImmune} immunization method, we design two random immunization methods, as well as several heuristic immunization methods as our baselines. Besides, we also compare our \textit{AdvImmune} immunization method with state-of-the-art defense method.

$\bullet$ \textbf{Random immunization methods.} We consider two random methods with different candidate sets of immune node pairs. \textit{Random} method selects immune node pairs randomly from all node pairs, while \textit{Attack Random} selects immune node pairs from the worst-case perturbed edges obtained by the surrogate model. Note that since \textit{Attack Random} knows the worst-case perturbed edges of surrogate model attacks, it is stronger than the \textit{Random} baseline.

$\bullet$ \textbf{Heuristic immunization methods.} 
 \emph{(1) Attribute-based methods.}
Researches show that attackers tend to remove edges between nodes from the same class and add edges to node pair from different classes~\cite{zugner_adversarial_2019,Wu2019AdversarialEF}. Therefore, we design baselines that maintain the connection of node pairs with high similarity between the attributes of nodes under the same class and the disconnection of node pair with low attribute similarity under different classes. We consider the commonly used Jaccard score and cosine score as the measure of similarity, namely \textit{Jaccard} and \textit{Cosine} respectively. 
\emph{(2) Structure-based methods.} We also design baselines considering the structure importance of edges. Specifically, for global structure importance, we use edge-betweenness as a measurement~\cite{Girvan2002CommunitySI}.
Locally, we adopt Jaccard similarity between the neighbors' labels of node pair. These two indicators reflect the connectivity and importance of edges, and we heuristically choose the immune node pairs with the greatest values, namely \textit{Betweenness} and \textit{Bridgeness} respectively.

$\bullet$ \textbf{Defense methods.} We use state-of-the-art defense method, robust training~\cite{Bojchevski2019CertifiableRT}, as our strong baseline. Robust training aims to improve the certifiable robustness of graph by retraining the GNN model for optimizing the \textit{robust cross-entropy loss} and \textit{robust hinge loss}, namely \textit{RCE} and \textit{CEM}.  Note that this is a strong baseline since it also devotes to improving certifiable robustness against any admissible attack. However, such robust training method suffers from the performance drop on clean graph and lacks flexibility to controlling the magnitude and effectiveness of defense.


\begin{figure*}[!h]
\centering
\subfigure[Ratio of robust nodes on Citeseer]{
\includegraphics[width=5.7cm]{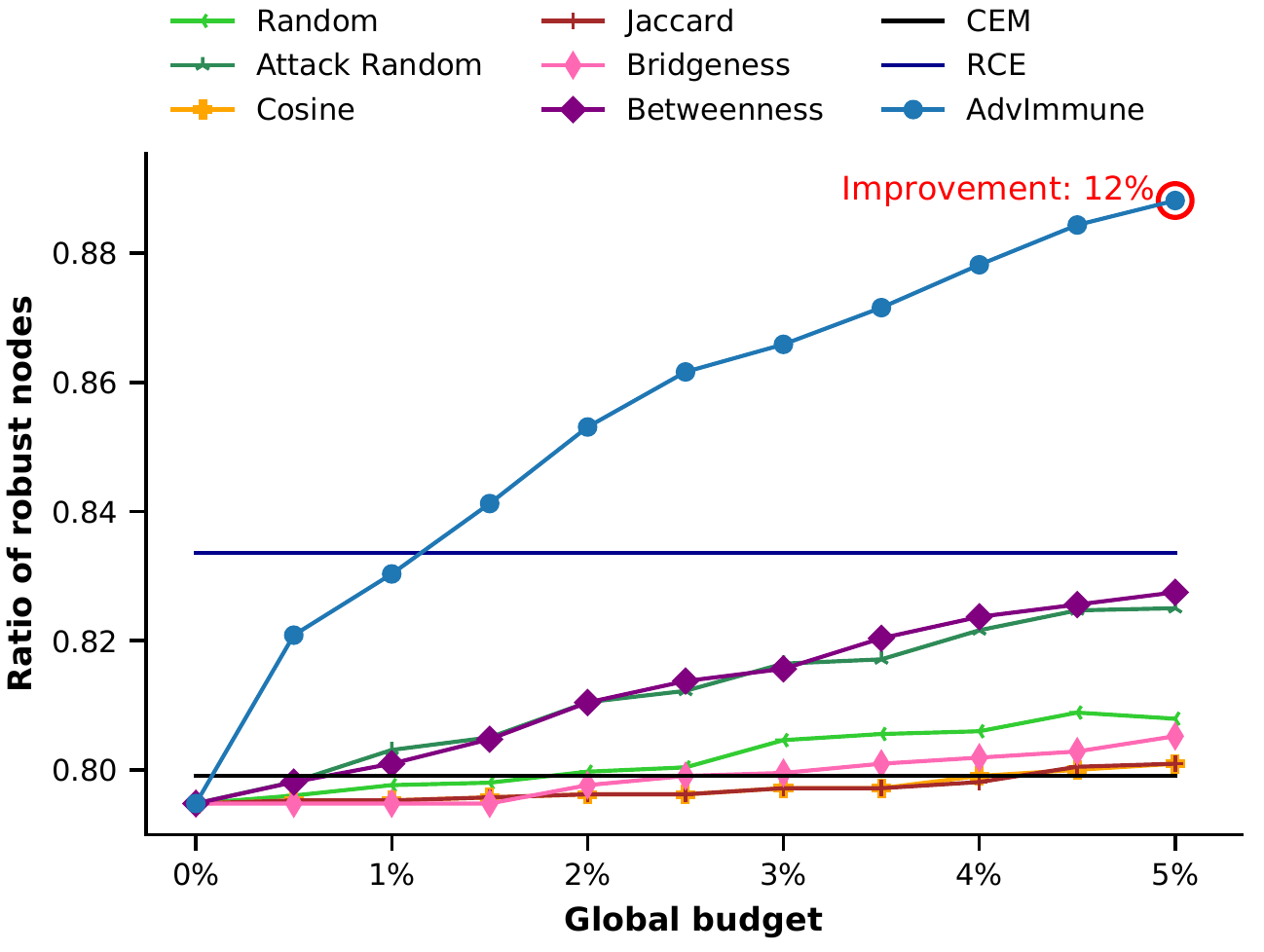}
\label{subfig:citeseer_r}
}
\subfigure[Ratio of robust nodes on Cora-ML]{
\includegraphics[width=5.7cm]{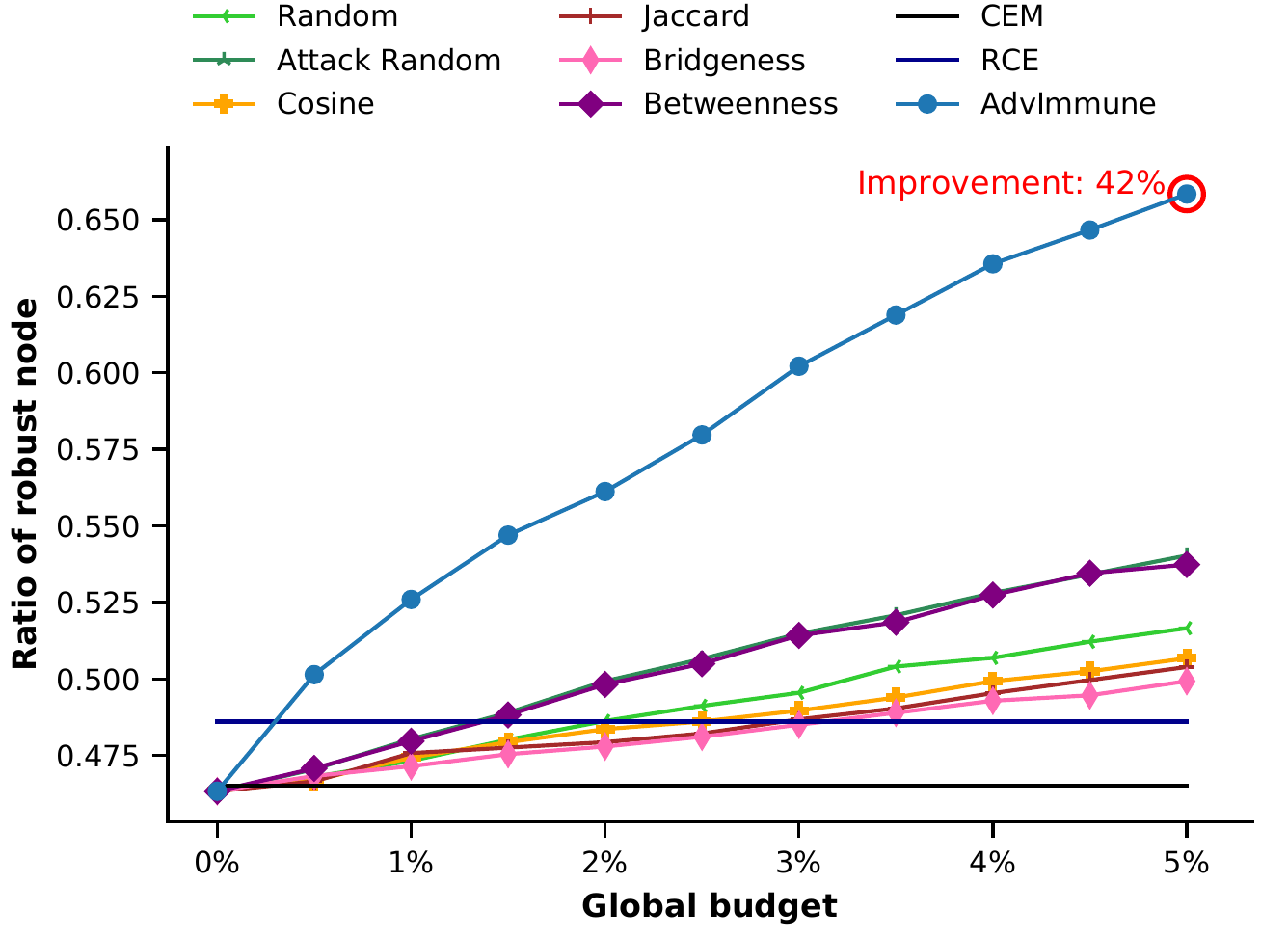}
\label{subfig:cora_r}
}
\subfigure[Ratio of robust nodes on Reddit]{
\includegraphics[width=5.7cm]{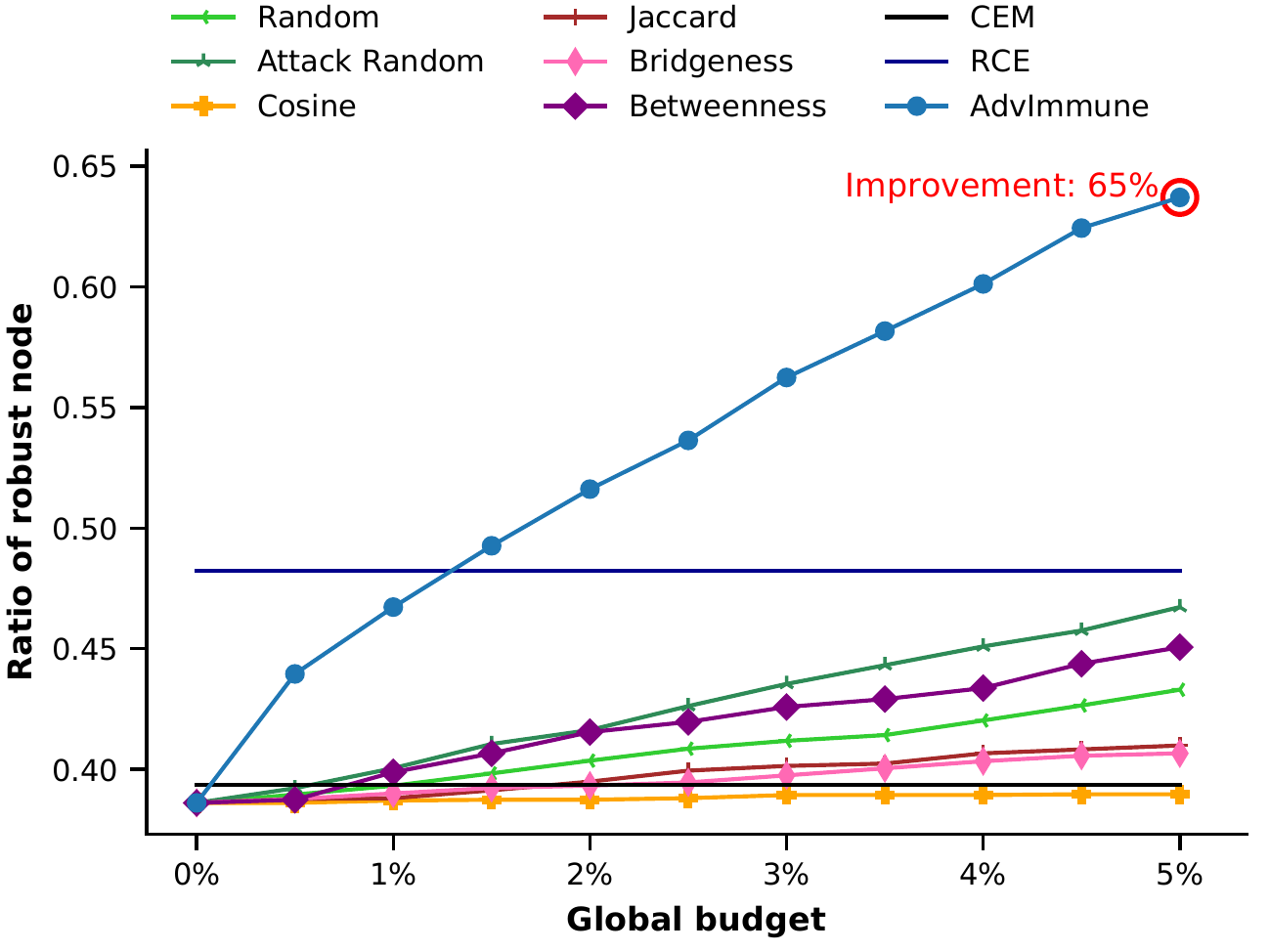}
\label{subfig:reddit_r}
}
\subfigure[Worst-case margin on Citeseer]{
\includegraphics[width=5.7cm]{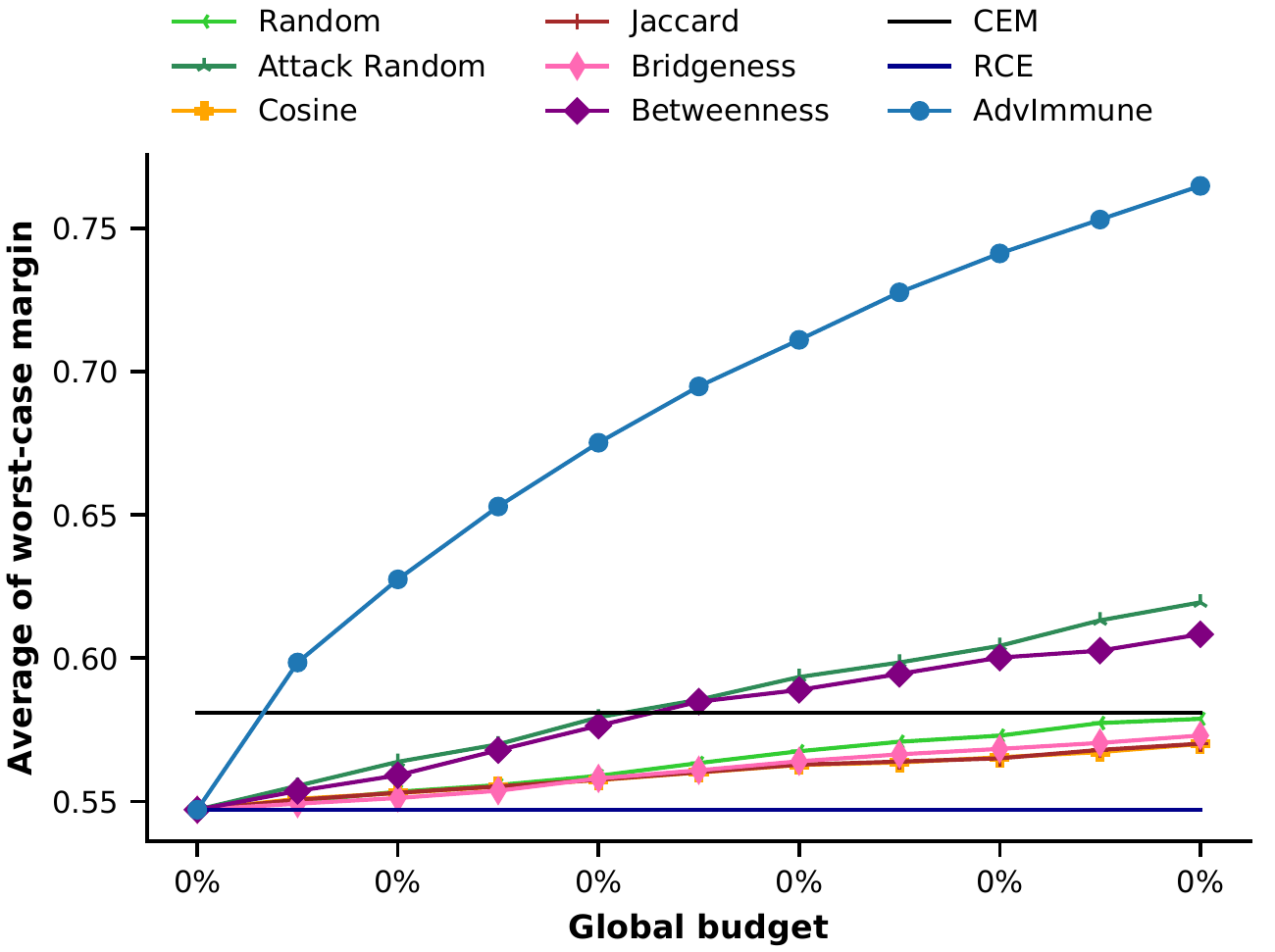}
\label{subfig:citeseer_w}
}
\subfigure[Worst-case margin on Cora-ML]{
\includegraphics[width=5.7cm]{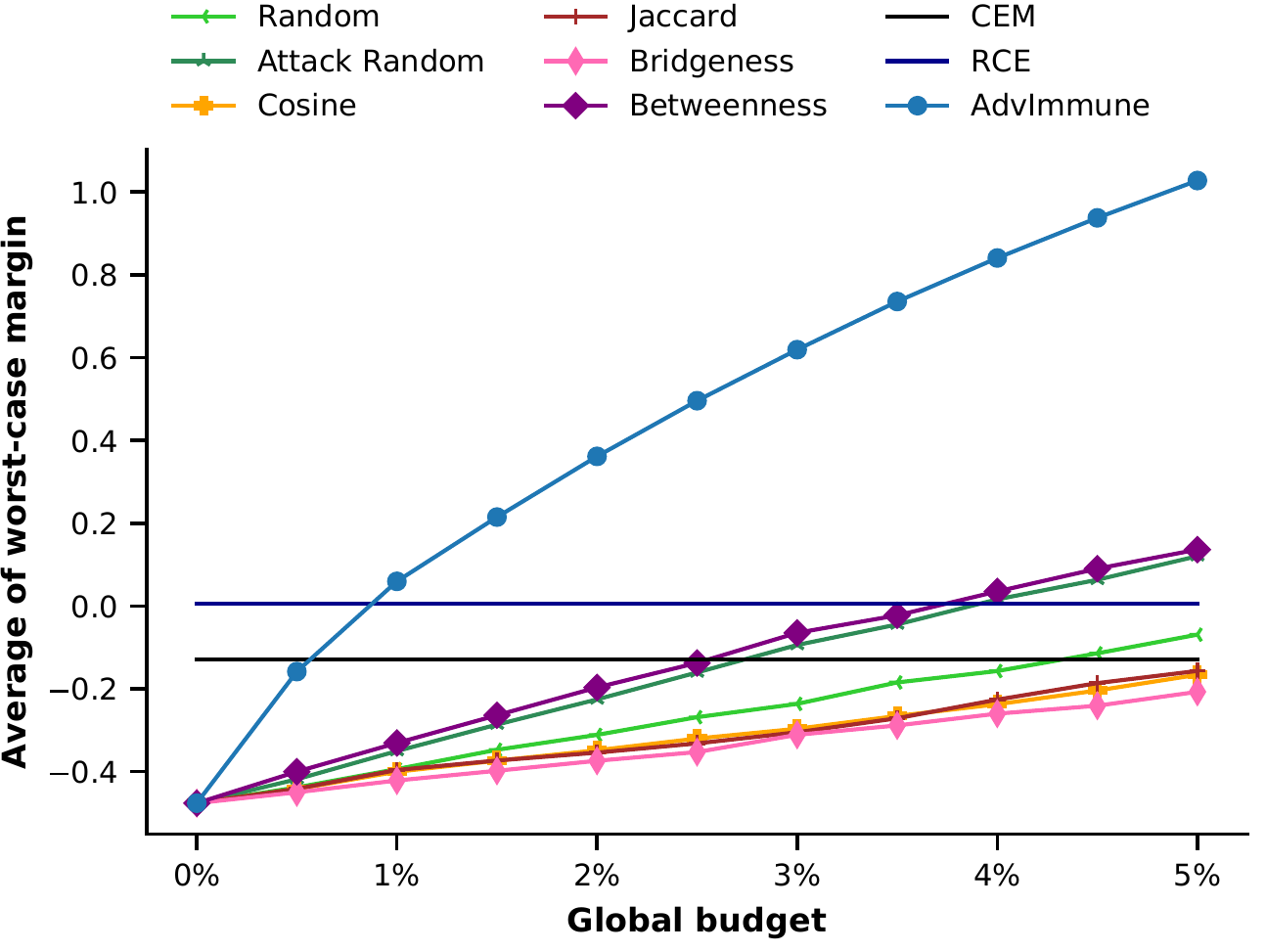}
\label{subfig:cora_w}
}
\subfigure[Worst-case margin on Reddit]{
\includegraphics[width=5.7cm]{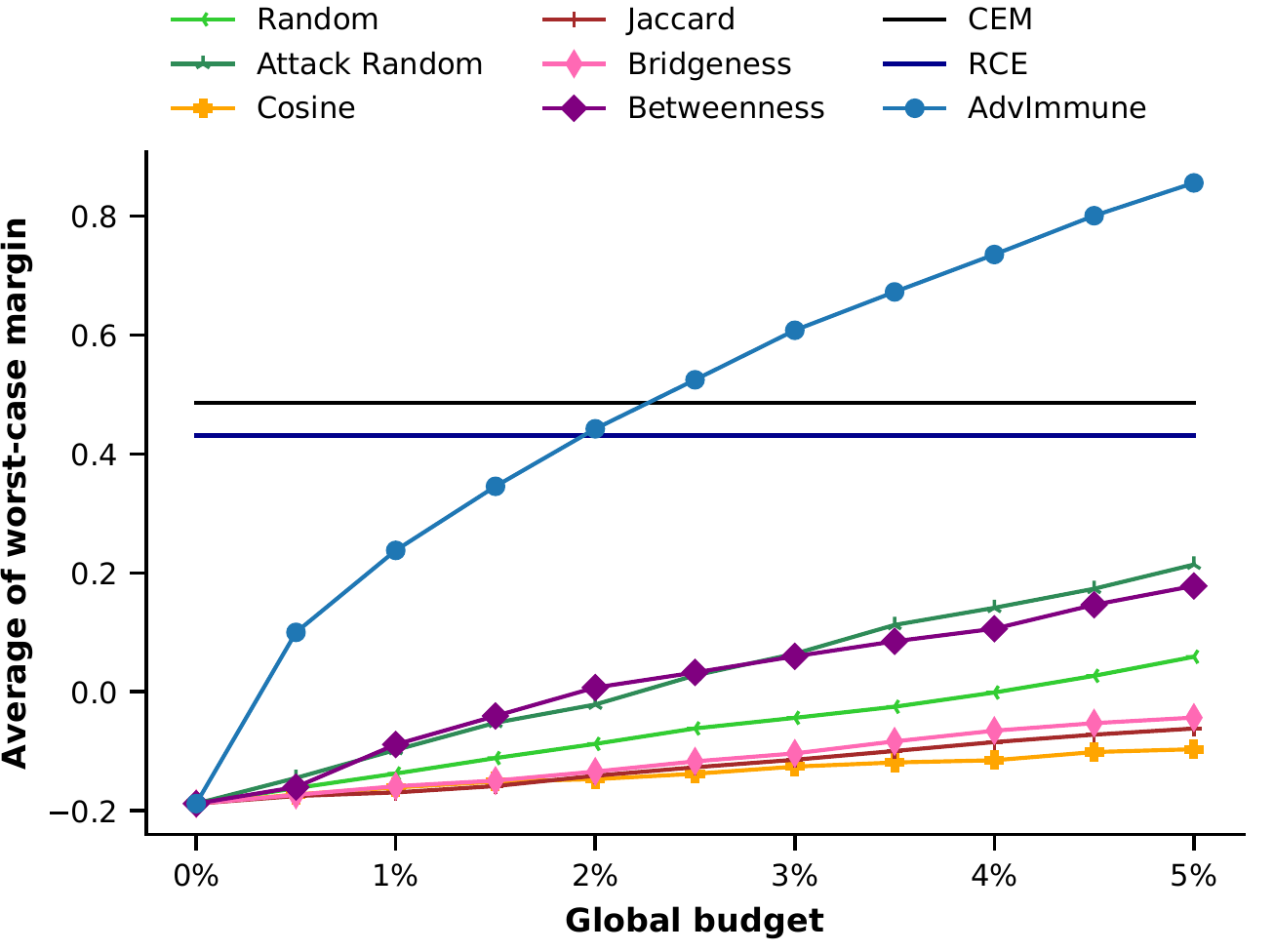}
\label{subfig:reddit_w}
}
\caption{Immunization performance in the scenario of \textbf{Remove-only}. The upper figures (a-c) show the changes of the ratio of robust nodes, while lower figures (d-f) show the changes of the average of worst-case margins.}
\label{fig:rem}
\end{figure*}

\subsection{Experimental Setup}
\label{subsec:scen}
The settings of GNN model in our experiments, i.e., $\pi$-PPNP, are the same with~\cite{Bojchevski2019CertifiableRT}. 
Specifically, we set the transition probability $\alpha=0.85$. 
As for our surrogate attack model, it has two settings of scenarios~\cite{Bojchevski2019CertifiableRT}. 
One is named as \textbf{Remove-only}, where $\mathcal{F}=\mathcal{E}\backslash \mathcal{E}_f$ for a given graph $G=(\mathcal{V}, \mathcal{E})$, i.e., attackers can only remove connected edges in graph. 
The other is named as \textbf{Remove-Add}, where the fragile edge set is $\mathcal{F}=\left(\mathcal{V} \times \mathcal{V}\right) \backslash \mathcal{E}_f$, i.e., the attacker is allowed to add or remove connection of any node pair in graph. 
Following the settings in robustness certification~\cite{Bojchevski2019CertifiableRT}, the fixed edge set is $\mathcal{E}_f=\mathcal{E}_{mst}$, where $\mathcal{E}_{mst}$ is edge set of the minimum spanning tree (MST) of graph, and the fragile edge is regarded as directed. 
For both surrogate model and robustness certification, the local budget is limited as $b_t=\boldsymbol{D}_t$ in \textbf{Remove-only} scenario and $b_t=\max\left(\boldsymbol{D}_t-6, 0\right)$ in \textbf{Remove-Add} scenario where $\boldsymbol{D}_t$ is the degree of node $t$. We do not limit global budget in both scenarios, i.e., $B=N^2$, in order to better demonstrate the effectiveness of our proposed \textit{AdvImmune} immunization even under strong attacks.
Following the settings of robustness certification~\cite{Bojchevski2019CertifiableRT,Zgner2019CertifiableRA}, we also use the prediction classes to measure robustness of nodes. 
Besides, policy iteration is adopted to generate the worst-case perturbation.

For \textit{AdvImmune} immunization, the immune \textit{local budget} is $c_t=\boldsymbol{D}_t$ in \textbf{Remove-only} and is not set in the other scenario. 
For baselines, random-based methods randomize 10 times in both scenarios. As for attribute-based methods, in \textbf{Remove-only} scenario, they only immunize the connected edges with high attribute similarity between nodes under the same class, while in \textbf{Remove-Add} scenario, they immunize connected edges with a probability of 0.3 and unconnected node pairs with a probability of 0.7. Such settings of probability is based on the fact that there are about 30\% removed edges among the worst-case perturbed edges in all datasets.

\subsection{Effectiveness of Adversarial Immunization}
\label{sec:main_exp}
To comprehensively demonstrate the effectiveness of our proposed \textit{AdvImmune} immunization, we conduct experiments on three datasets, i.e., Citeseer, Cora-ML, Reddit, and two certification scenarios, i.e., \textbf{Remove-only} and \textbf{Remove-Add}.

\subsubsection{Scenario of \textbf{Remove-only}}
In \textbf{Remove-only} scenario, the attacker can only remove connected edges. 
We show the performance from two aspects: the ratio of robust nodes and the average of worst-case margins. 
In Figure~\ref{fig:rem}, the upper figures show the changes of the ratio of robust nodes when varying global budget of immunization from 0.5$\%$ to 5$\%$ edges. and the lower figures show the changes of the average of worst-case margins.
Since only nodes with positive worst-case margin can be certified as robust, the sudden changes of certifiable robustness resulting in sudden jumps in the ratio of robust nodes. 
 As shown in Figure~\ref{fig:rem}, we can see that all immunization methods work better when global immune budget is increased, indicating that the more sufficient immune budget is, the higher certifiable robustness of graph is. 

For random-based immunization methods, \textit{Attack Random} performs better than \textit{Random} on all datasets. This is because \textit{Attack Random} knows the worst-case perturbed edges that the surrogate model attacks.
However, even with the same access of the worst-case perturbed graph, \textit{Attack Random} is far worse than our proposed \textit{AdvImmune} method, further proving the superiority of our method.

For heuristic-based immunization methods, both \textit{Jaccard} and \textit{Cosine} methods even perform worse than \textit{Random} method on three datasets. This result indicates that edges with high attribute similarity have less impact on certifiable robustness against any admissible attack, and may even bring some negative disturbance on the selection of immune node pairs.
\textit{Betweenness} is superior to \textit{Bridgeness}, indicating that global structure importance is more effective than local structure importance when selecting immune node pairs.

For state-of-the-art defense methods, \textit{RCE} performs better than \textit{CEM} on all datasets, which is consistent with the results in \cite{Bojchevski2019CertifiableRT}. However, \textit{AdvImmune} outperforms \textit{RCE} when only immunizing 0.5\% edges on Cora-ML, and immunizing 1.5\% edges on Citeseer and Reddit. Our proposed \textit{AdvImmune} method is very flexible, comparing to existing defense models that are fixed after training.

As for our proposed \textit{AdvImmune} immunization method, it significantly outperforms all the baselines on all datasets.
On Reddit (Figure~\ref{subfig:reddit_r},~\ref{subfig:reddit_w}), our \textit{AdvImmune} method increases the ratio of robust nodes by 65\% when immunizing only 5\% of all edges. 
On the other two datasets, Citeseer and Cora-ML, our method also brings 12\% and 42\% improvement, when immunizing 5\% edges. These results demonstrate the effectiveness of our proposed \textit{AdvImmune} immunization method, i.e., significantly improves the certifiable robustness of graph with an affordable immune budget.

\subsubsection{Scenario of \textbf{Remove-Add}}
We also conduct experiments in the scenario of \textbf{Remove-Add}, where the attacker is allowed to remove as well as add edges.
In this scenario, immune node pairs contain both connected edges and unconnected node pairs, and the global budget is set to be 1$\%$ of all node pairs. Note that the immunization in \textbf{Remove-Add} scenario is much difficult since the global budget of attack is $B=N^2$, which can thus better demonstrates the effectiveness of our proposed method. 
The baselines of structure-based immunization methods are ignored in this scenario since they can only calculate the importance of connected edges.

\begin{table}[htbp] 
 \caption{\label{tab:addrem} Immunization performance in the scenario of \textbf{Remove-Add}}
 \begin{tabular}{c|c|cc} 
  \toprule 
 Dataset & Methods & \tabincell{c}{Ratio of \\ robust nodes} & Improvement \\ 
  \midrule 
  \multirow{8}{*}{Citeseer} & No defense & 0.4806  & - \\ 
   & Random & 0.4807 & 0.02\%  \\ 
   & Attack Random & 0.5569   &  15.88\%\\
   & Jaccard & 0.4910  &  2.17\% \\ 
   & Cosine & 0.4910   & 2.17\% \\ 
   & CEM & 0.5246 & 9.17\%  \\ 
   & RCE & 0.5441   &  13.21\%\\
   & AdvImmune &\textbf{0.6336}  & \textbf{31.85\%} \\ 
   \midrule 
  \multirow{8}{*}{Cora-ML} & No defense & 0.1306  & - \\ 
   & Random &  0.1310 & 0.27\%  \\ 
   & Attack Random & 0.1867   &  42.92\%\\ 
   & Jaccard &  0.1552  &  18.80\% \\ 
   & Cosine & 0.1577   & 20.71\% \\ 
   & CEM & 0.1598 & 22.34\%  \\ 
   & RCE & 0.1335  &  2.18\%\\
   & AdvImmune &\textbf{0.2641}   & \textbf{102.18\%} \\
   \midrule 
  \multirow{8}{*}{Reddit} & No defense & 0.2727   & - \\ 
  & Random &  0.2729 & 0.06\%  \\ 
   & Attack Random &  0.3486 &  27.82\%\\
   & Jaccard & 0.2822  & 3.46\% \\ 
   & Cosine &  0.2822 & 3.46\% \\ 
   & CEM & 0.3747 & 35.76\%  \\ 
   & RCE & 0.3832 &  40.50\%\\ 
   & AdvImmune &\textbf{0.3982}   & \textbf{46.00\%} \\
  \bottomrule
 \end{tabular} 
\end{table}

Table ~\ref{tab:addrem} shows the ratio of robust nodes and the improvement percentage after immunization or defense.
The performance of immunization baselines are similar to that in the other scenario. 
\textit{Attack Random} performs better than \textit{Random}, while attribute-based heuristic immunization methods shows less effectiveness. As for defense baselines, i.e., \textit{RCE} and \textit{CEM}, they almost reach and perform the best comparing with heuristic immunization baselines. 

As for our proposed \emph{AdvImmune} method, it significantly outperforms all the baselines  on all datasets, achieving the improvement of robust nodes by 32\%, 46\%,102\%.
Especially on Cora-ML, our immunization method doubles the ratio of robust nodes. On Citeseer and Reddit, our method also improves the ratio of robust nodes remarkably, i.e., 31\% and 46\%. 
These results further demonstrate the effectiveness of our proposed \textit{AdvImmune} immunization method even in a difficult scenario.

\subsection{Transferability of Immunization under Various Attacks }

To verify the generality and transferability of the immune node pairs obtained by our proposed \emph{AdvImmune} method, we evaluate its  effectiveness against various attacks, i.e., state-of-the-art  \textit{metattack}~\cite{zugner_adversarial_2019} and \textit{surrogate attack}~\cite{Bojchevski2019CertifiableRT} obtained by the worst-case perturbed graph. 
Specifically, we first obtain immune node pairs by \textit{AdvImmune} and keep them unchanged the whole time. 
Then, we evaluate the transferability by node classification accuracy after defense or immunization on three settings, i.e., the clean graph, after \textit{surrogate attack}, and after \textit{metattack}, respectively.

We take robust training defense methods including \textit{CEM} and \textit{RCE} as our strong baselines.
Since \textit{metattack} uses GCN as a surrogate GNN model, we replace GCN with $\pi$-PPNP to obtain stronger attack.
Besides, we adopt the Meta-Self variant of \textit{metattack} since it is the most destructive one~\cite{Jin2020GraphSL}. 
As for the surrogate attack, it is obtained by the perturbed graph in robustness certification.
As for global budget, \textit{metattack} can perturb 20\% edges, while immune budget is only 5\% edges. The settings can better demonstrate the effectiveness of our proposed \textit{AdvImmune} method.

In Table ~\ref{tab:attack_acc}, we can see that on clean graph, defense methods (\textit{CEM}, \textit{RCE}) even lead to a decrease of classification accuracy, which is undesirable before attack actually happens.
As for our \textit{AdvImmune} method, it keeps the same accuracy as original GNN model, since it only immunizes edges without changing the graph or GNN model.
Under the setting of surrogate attack, both defense methods do not improve model performance after attack, when compared with the model performance with \textit{no defense}. This may due to the large decrease of accuracy on clean graph, making the defense less effective.
As for our method, the accuracy after immunizing only 5$\%$ edges outperforms all baselines and  brings significant improvement of model performance. 
The same immune node pairs can also significantly improve the model performance under \textit{metattack}.
These results prove that the immune node pairs are transferable against various attacks, improving the model performance under attacks while maintaining accuracy on the clean  graph.

\begin{table}[tbp] 
 \caption{\label{tab:attack_acc} The accuracy of node classification by GNN model under various attacks.} 
 \begin{tabular}{c|c|ccc} 
  \toprule 
 Dataset & Methods & \tabincell{c}{Clean \\graph}  & \tabincell{c}{Surrogate \\attack} & metattack \\ 
  \midrule 
  \multirow{4}{*}{Citeseer} & No defense & 0.74455 & 0.74313  & 0.72891 \\ 
   & CEM & 0.71896 &0.71137 & 0.67773 \\ 
   & RCE & 0.73555 &0.74313  & 0.72701 \\ 
   & AdvImmune &\textbf{0.74455}  & \textbf{0.74550} & \textbf{0.74265} \\ 
   \midrule 
  \multirow{4}{*}{Cora-ML} & No defense & 0.83701  & 0.79751  & 0.77473 \\ 
   & CEM & 0.80925  & 0.77651 & 0.71637 \\ 
   & RCE & 0.81957 & 0.77829  & 0.73416 \\ 
   & AdvImmune & \textbf{0.83701} & \textbf{0.83523} & \textbf{0.81459}\\ 
   \midrule 
  \multirow{4}{*}{Reddit} & No defense & 0.90453 &0.84816  & 0.79179 \\ 
   & CEM &0.86347 & 0.82926 & 0.77452\\ 
   & RCE & 0.86706& 0.84718 & 0.79700\\ 
   & AdvImmune & \textbf{0.90453} &\textbf{0.85598}  &\textbf{0.84001}\\
  \bottomrule
 \end{tabular}
\end{table}

\begin{figure}
\subfigure[Visualization of Immunization on Cora-ML]{
\includegraphics[width=6.24cm]{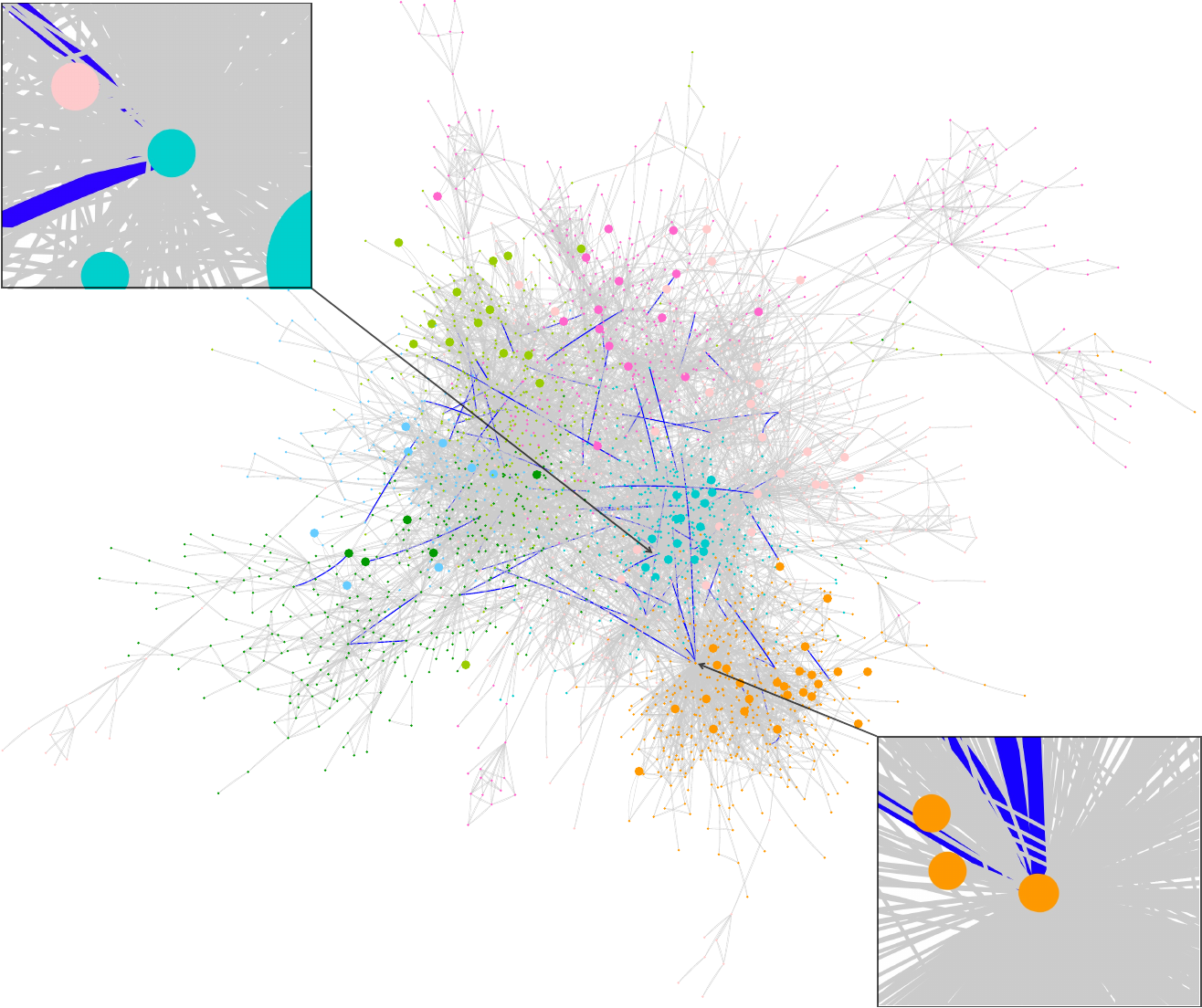}
\label{subfig:imm_cora}
}
\subfigure[Distribution of immune edges on different indicators]{
  \includegraphics[width=3.2cm]{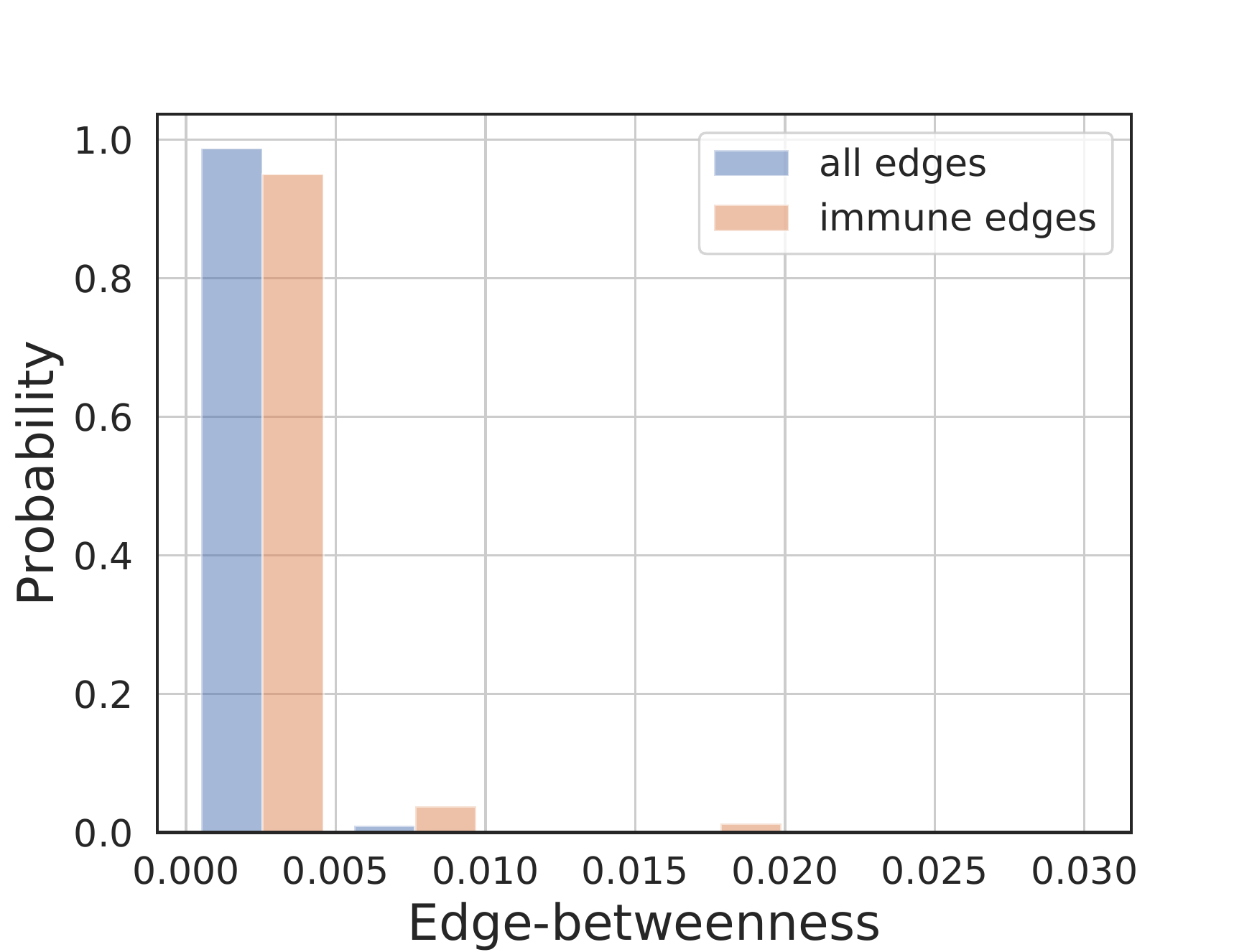}
 \includegraphics[width=2.98cm]{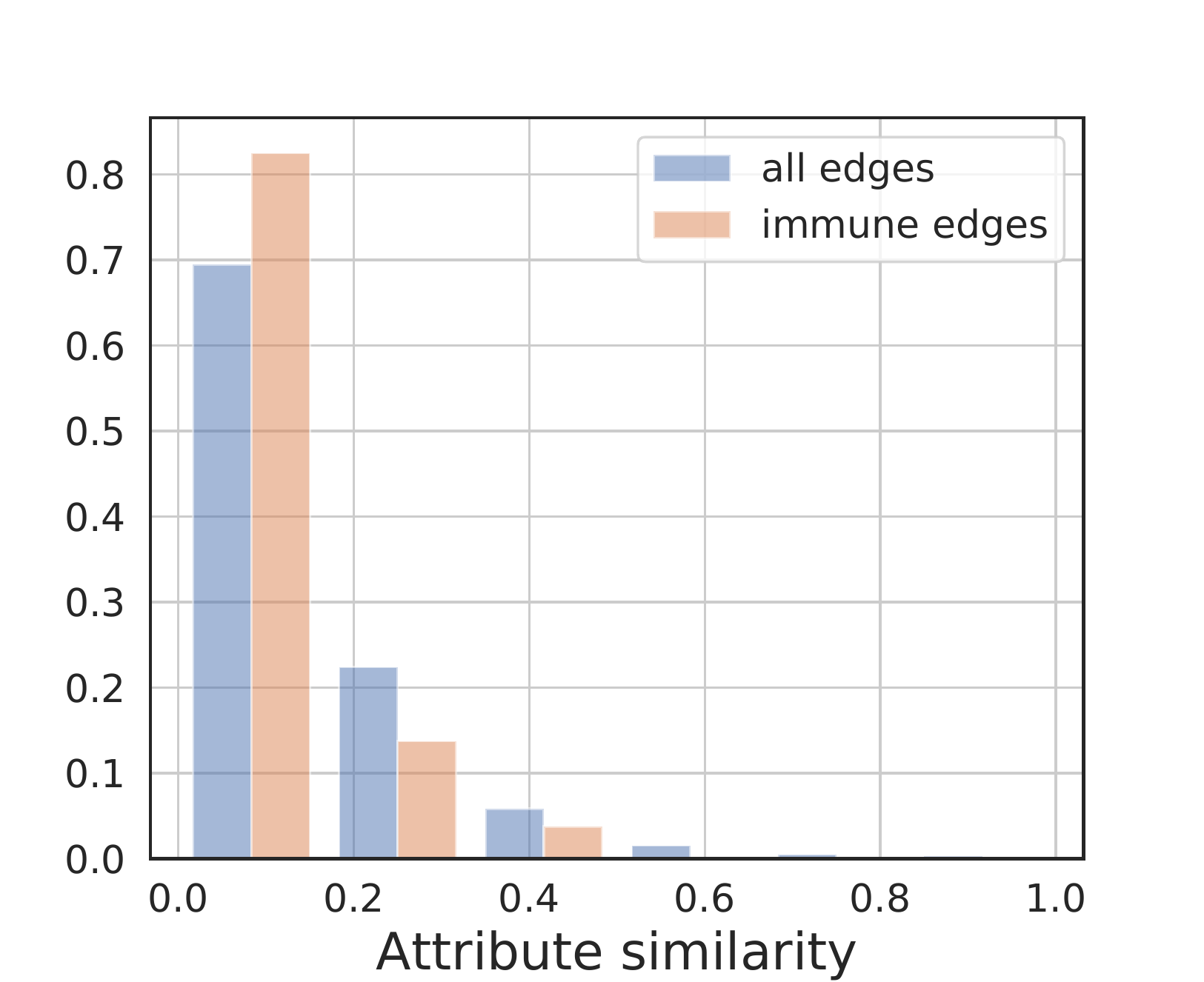}
\includegraphics[width=2.06cm]{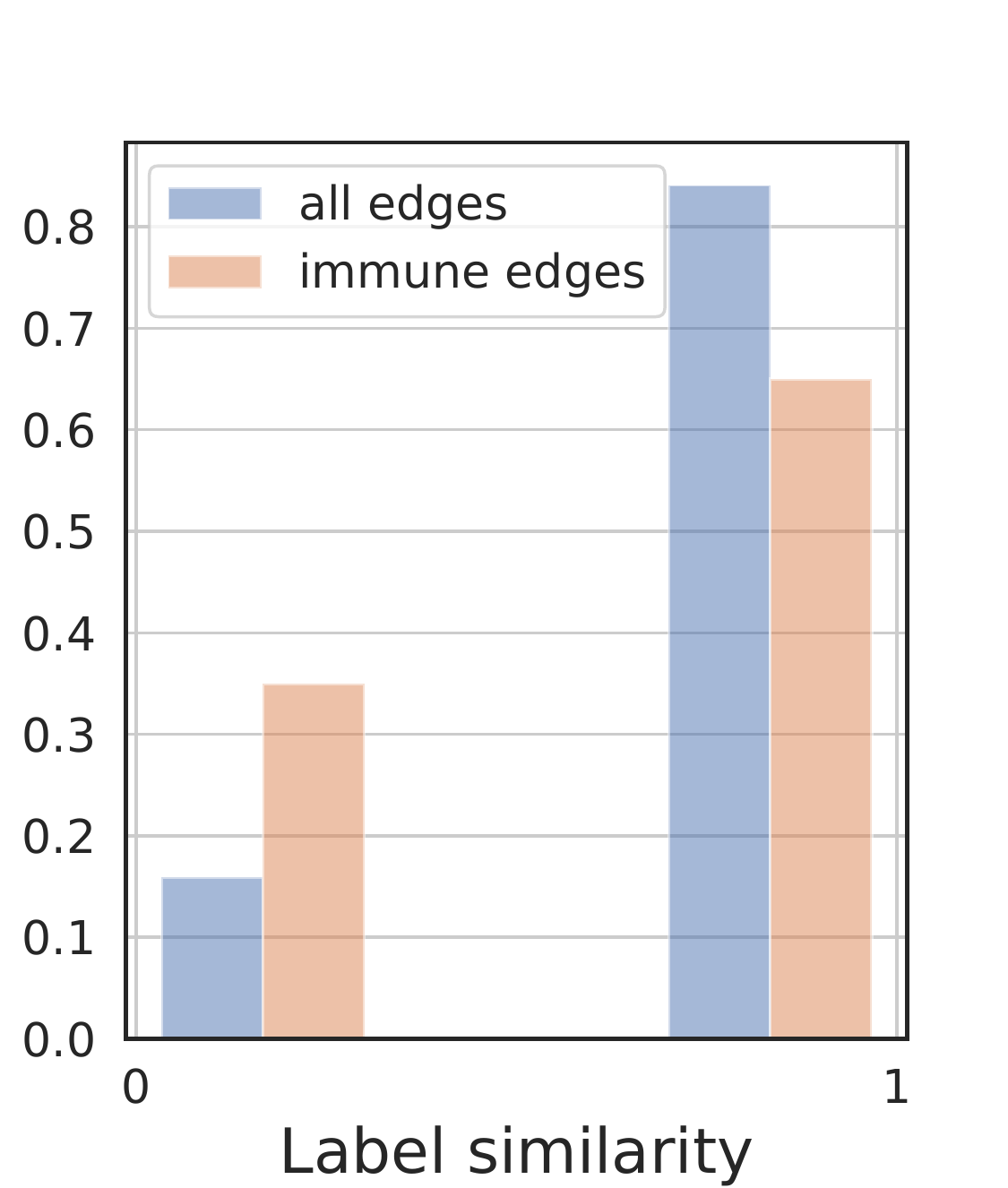}
 \label{subfig:dist}
}
\caption{(a) Visualization of Immunization on Cora-ML.
Larger nodes
 represent nodes that 
 become robust through immunization. 
 Insets enlarge the nodes connected to immune edges (blue).
(b) The distribution of immune edges on edge-betweenness, attribute similarity, and label similarity.
} 
\label{fig:case_study}
\end{figure}

\subsection{Case Study}
In order to have an intuitive understanding of our proposed \textit{AdvImmune} immunization method, we take Cora-ML as a case to study immune edges from three aspects, i.e., structure, attributes and labels. 
We take the scenario of \textbf{Remove-only} as an example,  and local immune budget is $c_t = \boldsymbol{D}_t$ and global budget is $C=5\%$ edges. Figure~\ref{subfig:imm_cora} visualizes the immune edges and the nodes that become robust through \textit{AdvImmune} immunization. 
Figure~\ref{subfig:dist} offers the distribution of immune edges on the above three aspects.

\textbf{Structure analysis.}
First, we take edge-betweenness as an indicator to analyze the structure characteristics of immune edges.
As shown in the left figure of Fig.~\ref{subfig:dist}, the edge-betweenness of immune edges is statistically larger than other edges.
This explains why the edge-betweenness immunization method is the strongest heuristic baseline.
In Figure~\ref{subfig:imm_cora}, the orange node with the highest degree, is the intersection of many immune edges. 
The other enlarged blue-green node connects with several immune edges, which also has high degree. 
This result indicates that \textit{AdvImmune}  immunization tends to immunize the edges with certain structural domination, which is critical for improving the certifiable robustness of graph.

\textbf{Attribute analysis.}
We analyze node attribute similarity of immune edges. 
In the middle figure of Fig.~\ref{subfig:dist},
the attribute similarity of nodes at both ends of immune edges is mainly distributed from 0 to 0.1, while that of all edges is distributed around 0.
The node attribute similarity of immune edges is lower.
This may be the reason why attribute-based baselines do not work well in Figure~\ref{fig:rem}.

\textbf{Label analysis.}
For labels, we analyze the nodes' label similarity of immune edges. 
There are about 80\% of all edges which have the same node label at both ends, while the ratio of immune edges is only 60\%.   
This may due to that personalized PageRank of $\pi$-PPNP: $\mathbf{\Pi} =(1-\alpha)\left(\boldsymbol{I}_{N}-\alpha \boldsymbol{D}^{-1} \boldsymbol{A} \right)^{-1}$ involves the inverse operation, making diffusion matrix be dense and heterogeneous edges be helpful by changing diffusion weight.
It's an interesting finding that immunizing such heterogeneous edges are also helpful for the certifiable robustness of graph, which may need further exploration. 

Experiments on Citeseer and Reddit give similar results which are not included because of space limitation.

\section{Related Work}
GNNs have shown exciting results on many graph mining tasks, e.g., node classification~\cite{Klicpera2018PredictTP,hamilton2017inductive}, network representation learning~\cite{10.1145/3097983.3098061}, graph classification~\cite{rieck2019persistent}. 
However, they are proved to be sensitive to adversarial attacks~\cite{Sun2018AdversarialAA}.
Attack methods can perturb both graph structure and node attributes~\cite{Chen2020ASO}, while structure-based attacks are more effective and result in more attention. 
Specifically, Nettack~\cite{zugner2018adversarial} attacks node attributes and graph structure with gradient.
RL-S2V~\cite{Dai2018AdversarialAO} uses reinforcement learning to flip edges.
Metattack~\cite{zugner_adversarial_2019} poisons graph structure with meta-gradient.
Others perturb structure with approximation techniques~\cite{Wang2019AttackingGC} or injecting nodes~\cite{Sun2020AdversarialAO,Wang2020ScalableAO}. 

Various defense methods are proposed against the above attacks~\cite{Jin2020AdversarialAA}.
RGCN~\cite{Zhu2019RobustGC} adopts Gaussian distributions as hidden representations, which can defend against nettack~\cite{zugner2018adversarial} and RL-S2V~\cite{Dai2018AdversarialAO}. 
Xu \textit{et al.}~\cite{Xu2019TopologyAA} present the adversarial training framework to defend against metattack~\cite{zugner_adversarial_2019}. 
Bayesian graph neural networks~\cite{Zhang2018BayesianGC}, trainable edge weights~\cite{Wu2019AdversarialEF}, transfer learning~\cite{Tang2020TransferringRF}, and preprocessing method~\cite{Entezari2020AllYN,Jin2020GraphSL} are also used in other defense methods. 


Throughout the development of graph adversarial learning, attack methods are always defended, and defense methods are also failed under the next attack.
This may lead to a cat-and-mouse game, limiting the development of graph adversarial learning.
Recently, robustness certification~\cite{Liu2020CertifiableRT} and robust training~\cite{bojchevski_sparsesmoothing_2020,Jia2020CertifiedRO} methods have appeared to fill this gap. Z\"ugner \textit{et al.}~\cite{Zgner2019CertifiableRA} verify certifiable robustness w.r.t. perturbations on attributes. 
Bojchevski \textit{et al.}~\cite{Bojchevski2019CertifiableRT} provide the certification w.r.t. perturbations on structures. 

Different from above researches, in this paper, 
we firstly explore the potential and practice of adversarial immunization for certifiable robustness of graph against any admissible adversarial attack.

\section{Conclusions}

In this paper, we firstly propose \emph{adversarial immunization} on graphs. 
From the perspective of graph structure,
adversarial immunization aims to improve the certifiable robustness of graph against any admissible attack by vaccinating a fraction of node pairs, connected or unconnected.
To circumvent the computationally expensive combinatorial optimization, we further propose an effective algorithm called \textit{AdvImmune}, which optimizes meta-gradient in a discrete way to figure out suitable immune node pairs.
The effectiveness of our proposed method are evaluated  on three well-known network datasets, including two citation networks and one social network. 
Experimental results reveal that our \textit{AdvImmune} approach significantly improves the certifiable robustness of graph. We also verify the generality and transferability of \textit{AdvImmune} under various attacks.
However, since our method is based on robustness certification, it can only be applied for PPNP-like GNN models. Also, the complexity of $\pi$-PPNP limits its application to large graphs. We tend to solve these challenges and explore the adversarial immunization focusing on both graph structures and node attributes in the future work.

\begin{acks}
This work is funded by the National Natural Science Foundation of China under grant numbers 62041207 and 91746301 and National Key R\&D Program of China (2020AAA0105200). This work is supported by Beijing Academy of Artificial Intelligence (BAAI). Huawei Shen is also funded by K.C. Wong Education Foundation.
\end{acks}

\clearpage
\bibliographystyle{ACM-Reference-Format}
\bibliography{wsdmfp655}


\end{document}